\documentclass[conference]{IEEEtran}
\IEEEoverridecommandlockouts
\newcommand \ignore[1]{}

\newcommand*{\affaddr}[1]{#1} % No op here. Customize it for different styles.
\newcommand*{\affmark}[1][*]{\textsuperscript{#1}}
\newcommand*{\email}[1]{\texttt{#1}}
\usepackage{url}
\usepackage{caption}
\usepackage{subcaption}
\usepackage{siunitx}
\usepackage{MnSymbol}

\usepackage{cite}
\usepackage{amsmath,amsfonts}
\usepackage{algorithmic}
\usepackage{graphicx}
\usepackage{textcomp}
\usepackage{xcolor}
\def\BibTeX{{\rm B\kern-.05em{\sc i\kern-.025em b}\kern-.08em
    T\kern-.1667em\lower.7ex\hbox{E}\kern-.125emX}}

\usepackage{fancyhdr}
\fancypagestyle{firststyle}{
  \fancyhf{}
  \fancyfoot[L]{\footnotesize \textcopyright\ 2026 IEEE. Personal use of this material is permitted. Permission from IEEE must be obtained for all other uses, in any current or future media, including reprinting/republishing this material for advertising or promotional purposes, creating new collective works, for resale or redistribution to servers or lists, or reuse of any copyrighted component of this work in other works.}
  
}

\begin{document}

\title{
%Quantum Leap or Quantum Creep? Benchmarking Classical vs. Quantum Convolutional Neural Networks on Efficiency and Security\\
%Quantum Leap or Creep? Benchmarking Classical vs. Quantum Machine Learning Models in Image Recognition\\
%Quantum Leap or Creep? Benchmarking Classical vs. Quantum Machine Learning Models\\
%Quantum Leap or Creep? Benchmarking Classical vs. Quantum Models in \textcolor{red}{Image Recognition}\\
%Quantum Leap or Creep? Benchmarking Classical vs. Quantum Models in Computer Vision\\
Do We Really Need Quantum Machine Learning?: A Multidimensional Empirical Study
\thanks{$^*$ Corresponding author. $^e$ equal contribution.}
}

% \and
% \and
% \ignore{=========
% % \and
% % \IEEEauthorblockN{4\textsuperscript{th} Given Name Surname}
% % \IEEEauthorblockA{\textit{dept. name of organization (of Aff.)} \\
% % \textit{name of organization (of Aff.)}\\
% % City, Country \\
% % email address or ORCID}
% % \and
% % \IEEEauthorblockN{5\textsuperscript{th} Given Name Surname}
% % \IEEEauthorblockA{\textit{dept. name of organization (of Aff.)} \\
% % \textit{name of organization (of Aff.)}\\
% % City, Country \\
% % email address or ORCID}
% % \and
% % \IEEEauthorblockN{6\textsuperscript{th} Given Name Surname}
% % \IEEEauthorblockA{\textit{dept. name of organization (of Aff.)} \\
% % \textit{name of organization (of Aff.)}\\
% % City, Country \\
% % email address or ORCID}
% =========}
% }
\author{Sudip Vhaduri$^e$\affmark[1], Ryan Gammon$^e$\affmark[2], and Sayanton Dibbo$^*$\affmark[1]\\
\affaddr{\affmark[1]Department of Computer Science, University of Alabama, AL 35487}\\
\email{\affmark[1]\{svhaduri,sdibbo}\}@ua.edu,
\affmark[2]rdgammon@crimson.ua.edu}

\maketitle
\thispagestyle{firststyle}

\begin{abstract}
% This document is a model and instructions for \LaTeX. This and the IEEEtran.cls file define the components of your paper [title, text, heads, etc.]. *CRITICAL: Do Not Use Symbols, Special Characters, Footnotes, or Math in the paper title or Abstract.
The rapid growth of computer vision and increasingly complex image recognition tasks has exposed fundamental computational limitations of classical machine learning models, motivating the exploration of quantum computing as an emerging new paradigm. This paper presents a comprehensive benchmarking study of classical and quantum machine learning models for image recognition on the MNIST handwritten digit dataset, evaluating both traditional models, a Classical Support Vector Machine (CSVM) and a Quantum Support Vector Machine (QSVM), and deep neural network models, a Classical Convolutional Neural Network (CCNN) and a Quantum Convolutional Neural Network (QCNN), across four performance dimensions: classification accuracy, computational runtime, parameter count, and memory requirements. Experiments are conducted as functions of both feature dimensionality and sample size, and across CPU and GPU execution environments, providing a controlled, multidimensional comparison to address gaps in prior work. For the SVM-based models, QSVM consistently outperforms CSVM in accuracy, reaching $\sim$ 0.90 versus $\sim$ 0.85 at 1,000 samples, with a higher computational cost. A feature count of 10 qubits and a sample size in the range of 200 -- 500 emerge as practical operating points that balance accuracy and runtime. For the neural network models, CCNN and QCNN achieve comparable classification accuracy, both exceeding 0.96 at 64 features and 60,000 samples, yet QCNN offers superior parameter and memory efficiency at higher feature counts, while incurring higher runtime. 
Across both model families, quantum models consistently outperform classical models by greater margins in accuracy as feature dimensionality or sample size increases, suggesting that quantum models offer the greatest relative advantage in high-data or high-dimensionality regimes. 
These findings provide actionable guidance for practitioners considering quantum machine learning for computer vision tasks under real-world resource constraints, and directly inform the design of efficient, quantum-ready perception systems for the automated, connected, and safe transportation applications central to the mission of the Alabama Transportation Institute (ATI) and the National Center for Transportation Cybersecurity and Resiliency (TraCR).
\end{abstract}


\section{Introduction}\label{sec:intro}
% This document is a model and instructions for \LaTeX.
% Please observe the conference page limits~\cite{li2024membership}. 

Computer vision and image recognition have become indispensable pillars of modern artificial intelligence, underpinning transformative applications in autonomous vehicles, medical diagnostics, industrial quality control, and surveillance~\cite{KAUR2023103812,vhaduri2023implicit, dibbo2024improving,vhaduri2023environment,amebley2025neuro,vhaduri2023bag}. The global computer vision market was valued at approximately \$19.82 billion in 2024 and is projected to reach \$58.29 billion by 2030, growing at a compound annual growth rate (CAGR) of 19.8\%~\cite{grandview_cv_2024}. Classical machine learning paradigms, most prominently Support Vector Machines (SVMs) and Convolutional Neural Networks (CNNs), have long driven progress in image classification tasks~\cite{kharsa2023advances,peral2024systematic}. However, as visual datasets scale into the hundreds of millions and recognition tasks grow increasingly complex, these classical approaches confront fundamental computational bottlenecks: SVMs suffer from complex kernel construction, while deep CNNs demand hundreds of millions of parameters and substantial memory footprints, rendering large-scale deployment computationally prohibitive~\cite{kharsa2023advances}. Such constraints have motivated the search for radically different computing substrates capable of representing data more efficiently and capturing complex feature interactions that challenge classical models.

Quantum computing has emerged as a compelling new paradigm, with the global quantum computing market estimated at \$1.42 billion in 2024 and projected to reach \$4.24 billion by 2030 at a CAGR of 20.5\%~\cite{grandview_qc_2024}. Quantum machine learning (QML) exploits the principles of superposition and entanglement to represent data in exponentially high-dimensional Hilbert spaces, offering the potential for fundamentally more expressive feature maps than their classical counterparts~\cite{peral2024systematic,Havl_ek_2019,10.3389/fams.2021.716044}. 
\ignore{Pioneering architectures such as the Quantum Convolutional Neural Network (QCNN)~\cite{cong2019quantum} and quantum-enhanced kernel methods have demonstrated promising results for image recognition, with hybrid quantum-classical models achieving over 99\% classification accuracy on Medical MNIST and over 82\% on CIFAR-10~\cite{senokosov2024quantum}. 
Despite this momentum, a critical gap persists in the literature: existing studies evaluate quantum models narrowly, focusing on accuracy or at most one secondary metric, without situating quantum and classical models within a rigorous, \textit{simultaneous} multi-dimensional evaluation framework spanning accuracy, runtime, parameter efficiency, and memory requirements as joint functions of both feature dimensionality and sample size~\cite{kharsa2023advances, peral2024systematic}. Moreover, no prior study has systematically compared both traditional machine learning models (i.e., SVMs) and deep neural network models (i.e., CNNs) under identical experimental conditions across CPU and GPU execution environments. This work addresses these gaps through a comprehensive benchmarking study of classical and quantum SVMs and CNNs on the MNIST dataset, yielding actionable insights into the regimes, in terms of data scale and dimensionality, where quantum models offer a genuine and measurable advantage over their classical counterparts.}

% Add as the final paragraph of the Introduction: 
This work is motivated in part by the applied transportation research mission of the Alabama Transportation Institute (ATI) at The University of Alabama, which is committed to achieving \textit{ACES\textsuperscript{2} Mobility}--- Automated, Connected, Electric, Shared, and Safe transportation---for the residents of Alabama and beyond~\cite{ati_mission}. Computer vision is a foundational enabling technology for this vision: autonomous perception, real-time traffic monitoring, and vehicle classification in connected and autonomous vehicles (CAVs) all rely critically on fast, accurate, and resource-efficient image recognition models. Additionally, this work aligns with the mission of the National Center for Transportation Cybersecurity and Resiliency (TraCR), headquartered at Clemson University, which identifies quantum computing and artificial intelligence as key technologies for hardening transportation cyber-physical systems against emerging threats~\cite{tracr_mission}. By systematically benchmarking classical and quantum machine learning models across accuracy, runtime, parameter count, and memory, this paper aims to provide the transportation research community with concrete empirical guidance on whether, and under what conditions, quantum models offer a practical advantage for the vision and cybersecurity workloads of next-generation intelligent transportation systems.

% \section{Preliminaries}\label{preli}
\section{Related Work}\label{sec:preli}

Due to the ever-growing importance of computer vision and image recognition/classification, researchers have been using various classical machine learning and deep learning modeling techniques, such as support vector machines, convolutional neural networks, and others~\cite{KAUR2023103812,vhaduri2023implicit, dibbo2024improving,vhaduri2023environment,amebley2025neuro,vhaduri2023bag
}. As recognition tasks become more complex and require multidimensional improvements, the classical techniques struggle due to their inherent shortcomings. Therefore, researchers have started exploring the recent emerging computing paradigm, i.e., quantum machine learning, in computer vision and image recognition, like other areas~\cite{dibbo2016efficient}, due to its ability to represent data in high-dimensional Hilbert spaces~\cite{Havl_ek_2019} and capture complex interactions through quantum entanglement~\cite{10.3389/fams.2021.716044}. 

In this new era of quantum computing, while some efforts have been made to develop a quantum computing-based generative model of synthetic image generation~\cite{nokhwal2023quantumgenerativeadversarialnetworks}, image recognition has received limited attention, with some scattered and independent efforts being made either to achieve a better accuracy with fewer parameters in deep neural networks, such as the convolutional neural networks~\cite{senokosov2024quantum,cong2019quantum}, to reduce quantum bit storage in traditional machine learning (ML) models, such as the support vector machines~\cite{Parigi_2026}, or simply to achieve a higher accuracy~\cite{DBLP:journals/corr/abs-1805-06260}. Yet other critical evaluation metrics, such as computational runtime and scalability, often received little attention.

However, while these quantum machine learning works primarily focus on the accuracy or at most one secondary metric, like the number of parameters or quantum bit storage for either traditional ML models or deep neural network models, current research lacks a comprehensive comparison between quantum and classical image recognition using both traditional machine learning and deep neural network models across multidimensional evaluations simultaneously. In particular, we found no studies that jointly analyze accuracy, runtime, parameter efficiency, and memory/storage requirements as functions of either feature dimensionality or sample size, under comparable experimental conditions such as CPU or GPU. {\bf Our work aims} to address these gaps by systematically evaluating these metrics as joint functions of sample size and feature dimensionality across both traditional machine learning models, such as SVMs, and deep neural network models, such as CNNs, and by additionally comparing performance across CPU and GPU execution environments.
\section{Methods}\label{methods}
In this section, we present the MNIST dataset used in this paper (Section~\ref{dataset}), the set of machine learning models (Section~\ref{models}), and our experimental setup (Section~\ref{exp}). Before presenting the details, we will outline the performance measures used in this work.

\subsection{\textcolor{black}{Preliminaries: Performance Metrics}}\label{perMeasures}

In this work, we consider the following metrics to evaluate the performance of different modeling schemes/setups:

{\bf {\em Accuracy}} is the fraction of correct classification/identification, defined as below, 
\begin{equation}
\label{acc}
% ACC = (TP+TN)*(TP+FN+FP+TN)^{-1}
% ACC = \frac{TP+TN}{TP+FN+FP+TN}
Accuracy = \frac{\sum_{i=1}^{N} c_i
}{N}
\end{equation}

Where,
$$c_i = \begin{cases}
1 & \text{if $i^{th}$ classification is correct} \\
0 & \text{otherwise}
\end{cases}$$

And, $N$ is the total number of classifications in a multi-class classification task. 
We also use {\bf {\em runtime}}, in seconds (s), to compare the performance of models. Further, for the deep neural networks, we consider the total number of {\bf{\em parameters}} and the {\bf {\em memory}} requirements in kilobytes (kB).

% \subsection{Dataset}\label{dataset}
%\subsection{Image Recognition Dataset}\label{dataset}
\subsection{Image Dataset}\label{dataset}
The experiments were conducted using the MNIST\cite{726791} handwritten digit dataset, accessed through the \texttt{fetch\_openml} function from scikit-learn. MNIST contains 70,000 grayscale handwritten digits labeled 0 -- 9. In our experiment, we use scikit-learn's \texttt{train\_test\_split} to use 75\% for training and 25\% for testing. The original dimensionality of these grayscale images is 28 x 28 = 784. The pixel values are normalized, resulting in each having a value of 0 to 1, representing how shaded the region is.

\subsection{Machine Learning Models}\label{models}

In this paper, we experiment with traditional and quantum machine learning models described below. 

% \subsubsection{\textcolor{red}{Support Vector Machine (SVM)}}
% For our classical Support Vector Machine (CSVM), we utilize scikit-learn's SVM module, as well as their PCA module for feature reduction. Following PCA reduction, features were standardized using scikit-learn's StandardScaler method. The kernel used was cosine similarity, as we recognized its similarity to the high-dimensional embedding of the Quantum SVM. We measured runtime and accuracy by changing sample size as the independent variable, followed by feature dimensions given by PCA. The configurations were a sample list of [50, 100, 200, 500, 1000] with 12 features, and a feature list of [2, 4, 6, 8, 10, 12] with samples fixed at 300.

\subsubsection{Classical Support Vector Machine (SVM) Baseline}
For the classical SVM (CSVM) baseline, we use the Support Vector Machine implementation provided by \texttt{scikit-learn} (\texttt{SVC}). The MNIST dataset is first truncated to a fixed number of samples and reduced in dimensionality using Principal Component Analysis (PCA), where the number of principal components is treated as a tunable hyperparameter.
After dimensionality reduction, the features are standardized using \texttt{StandardScaler}. The dataset is then split into training and testing sets \ignore{using a 75\%:25\% ratio }with a fixed random seed for reproducibility.

We evaluate a Support Vector Machine using a precomputed cosine similarity kernel. Specifically, the kernel matrix is explicitly constructed using pairwise cosine similarities between feature vectors and passed to the SVM as a precomputed Gram matrix.
Cosine similarity between two feature vectors $\mathbf{A}$ and $\mathbf{B}$ is defined as:
\[
\cos(\theta) = \frac{\mathbf{A} \cdot \mathbf{B}}{\|\mathbf{A}\|\|\mathbf{B}\|}
\]

Using this measure, we construct the Gram matrix $G \in \mathbb{R}^{n \times n}$, where each entry corresponds to the cosine similarity between two samples:
\[
G_{ij} = \cos(x_i, x_j)
\]
\[
G =
\begin{bmatrix}
\cos(x_1, x_1) & \cos(x_1, x_2) & \cdots & \cos(x_1, x_n) \\
\cos(x_2, x_1) & \cos(x_2, x_2) & \cdots & \cos(x_2, x_n) \\
\vdots & \vdots & \ddots & \vdots \\
\cos(x_n, x_1) & \cos(x_n, x_2) & \cdots & \cos(x_n, x_n)
\end{bmatrix}
\]

% Two experimental settings are considered:
% \begin{enumerate}
%     \item \textbf{Feature dimension scaling:} the number of PCA components is varied in \{2, 4, 6, 8, 10, 12\}, while the number of samples is fixed at 300.
%     \item \textbf{Sample size scaling:} the number of training samples is varied in \{50, 100, 200, 500, 1000\}, while the feature dimension is fixed at 12.
% \end{enumerate}
% In both settings, runtime and classification accuracy are recorded. The PCA dimension is aligned with the dimensionality used in the quantum model, enabling a controlled comparison between the classical SVM baseline and the quantum model.

% \subsubsection{\textcolor{red}{Quantum Support Vector Machine (QSVM)}}
% The quantum Support Vector Machine (QSVM) used the same techniques and standardization as the SVM, as well as the lists for experimentation, with the important note that the number of features is equivalent to the number of qubits used in the quantum embedding circuit. This ensured that each classical feature vector matched the dimensionality required for quantum angle embedding. Classical features were encoded into quantum states using angle embedding implemented through quantum simulation in PennyLane. The quantum kernel was constructed by computing the overlap between quantum states. 

\subsubsection{Quantum Support Vector Machine (QSVM)}
The Quantum Support Vector Machine (QSVM) uses the same preprocessing pipeline and experimental configurations as the classical SVM (CSVM), including PCA-based dimensionality reduction and standardization. The number of PCA components is set equal to the number of qubits in the quantum model, ensuring that each classical feature vector matches the dimensionality required for quantum encoding.

Formally, after dimensionality reduction, each input sample is represented as:
\[
x_i \in \mathbb{R}^d, \quad d = n_{\text{qubits}}
\]

Classical feature vectors are encoded into quantum states using an angle embedding function implemented in \texttt{PennyLane}. This embedding maps classical data to a parameterized quantum state:
\[
x_i \mapsto |\psi(x_i)\rangle
\]

The quantum feature map is realized through rotation-based encoding, where each feature is encoded into qubit rotation angles.

The quantum kernel is then defined as the overlap between quantum states:
\[
K(x_i, x_j) = \left|\langle \psi(x_i) | \psi(x_j) \rangle \right|^2
\]

This kernel is used to construct the Gram matrix:
\[
K_{ij} = \left|\langle \psi(x_i) | \psi(x_j) \rangle \right|^2
\]

which is passed to the SVM for classification in the same manner as the classical kernel.

We use the cosine similarity kernel as a classical baseline for comparison with the quantum kernel. Both approaches are based on inner products between feature representations; however, cosine similarity operates in a classical Euclidean feature space, whereas the quantum kernel is defined as the inner product between quantum states in a high-dimensional Hilbert space.

% \subsubsection{\textcolor{red}{Convolutional Neural Network (CNN)}}
% This study implemented a fully classical convolutional neural network (CCNN) architecture using PyTorch to provide a direct comparison against the hybrid quantum neural network (HQNN) framework. Data processing consisted of loading the images, converting them to tensors, splitting them into training and testing groups, and batching using data loaders. Similar to PCA in the SVM, feature reduction was introduced before classification. The dimensional reduction operation was implemented using a fully connected linear projection. The data was fed through three layers with ReLU activation before the final layer, which predicted the class of the digit. Our loss function was cross-entropy and the Adam optimizer was used for optimization. The learning rate selected was 0.001. The variables changed for testing were the feature numbers resulting from compression and sample size, the same as the SVM setup. The list of input variables is [16, 32, 64] and the training size list is [15,000, 30,000, 60,000].

\subsubsection{Classical Convolutional Neural Network (CNN) Baseline}
We implement a fully classical CNN (CCNN) in PyTorch to provide a baseline for comparison with the quantum neural network. 
The model follows a bottleneck architecture designed to match the dimensionality of the quantum model for a fair comparison.

Let each input image be denoted as:
\[
x_i \in \mathbb{R}^{28 \times 28}
\]

The first stage performs feature compression by flattening the image and applying a learned linear projection:
\[
z_i = W_r x_i + b_r, \quad z_i \in \mathbb{R}^d
\]
where \(d \in \{16, 32, 64\}\) is the reduced feature dimension used for comparison with the quantum CNN.

This dimensionality reduction layer plays the same role as PCA in the classical SVM baseline, but is learned directly through backpropagation.

The compressed features are then passed through a fully connected neural network:
\[
h_i^{(1)} = \sigma(W_1 z_i + b_1)
\]
\[
h_i^{(2)} = \sigma(W_2 h_i^{(1)} + b_2)
\]
where \(\sigma(\cdot)\) denotes the ReLU activation function.

Finally, the output layer produces class logits:
\[
\hat{y}_i = W_o h_i^{(2)} + b_o
\]

The model is trained using the cross-entropy loss:
\[
\mathcal{L} = - \sum_{i} y_i \log(\hat{y}_i)
\]

% Optimization is performed using the Adam optimizer with learning rate \( \eta = 0.001 \). Five epochs are used due to the large computational cost associated with training a quantum neural network.

% Two experimental variables are considered:
% \begin{enumerate}
%     \item \textbf{Feature dimension:} \( d \in \{16, 32, 64\} \), corresponding to the compressed representation size.
%     \item \textbf{Training set size:} \( N \in \{15{,}000, 30{,}000, 60{,}000\} \), used to evaluate scaling behavior.
% \end{enumerate}
% This setup mirrors the experimental configuration of the SVM and QSVM baselines to ensure a controlled comparison between quantum and classical SVMs and neural networks.

% \subsubsection{\textcolor{red}{Quantum Convolutional Neural Network (QCNN)}}
% Our quantum convolutional neural network (QCNN) setup mirrors the classical version, with the important distinction of the neural network layers being replaced with quantum layers using angle encoding, which contain trainable rotations. 4 qubits are used per circuit and the number of layers is equal to the number of features divided by 4. The quantum circuit outputs expectation values of Pauli-Z measurements. 

% \subsubsection{Quantum Convolutional Neural Network (HQNN)}
% The hybrid quantum neural network (HQNN) mirrors the architecture of the classical model, with the key distinction that the classical hidden layers are replaced by variational quantum circuits. 
\subsubsection{Quantum Convolutional Neural Network (QCNN)}
The quantum convolutional neural network (QCNN) mirrors the architecture of the classical model, with the key distinction that the classical hidden layers are replaced by variational quantum circuits. 
This enables a quantum feature transformation of classical data prior to classification.

Let each input image be represented as:
\[
x_i \in \mathbb{R}^{28 \times 28}
\]

A classical linear projection is first applied to reduce the dimensionality of the input:
\[
z_i = W_r x_i + b_r, \quad z_i \in \mathbb{R}^{n_{\text{features}}}
\]

The reduced feature vector is then partitioned into \(c\) quantum circuits, where:
\[
c = \frac{n_{\text{features}}}{q}
\]
and \(q = 4\) denotes the number of qubits per circuit.

Each feature vector is reshaped as:
\[
z_i \rightarrow \{z_i^{(1)}, z_i^{(2)}, \dots, z_i^{(c)}\}, \quad z_i^{(k)} \in \mathbb{R}^{q}
\]

Each sub-vector is encoded into a quantum state using angle embedding:
\[
z_i^{(k)} \mapsto |\psi(z_i^{(k)})\rangle
\]

The encoding is implemented using single-qubit rotations:
\[
R_X(\theta), \quad R_Y(\theta)
\]

where the input features are used as rotation angles.

Variational Quantum Circuit:
Each quantum circuit consists of a parameterized variational layer with trainable weights:
\[
U(\theta) = \prod_{l=1}^{D} U_l(\theta_l)
\]

where \(D\) is the circuit depth. Each layer applies:
- single-qubit rotations
- entangling CNOT gates

After applying the variational circuit, the output is measured using Pauli-Z expectation values:
\[
f(z_i^{(k)}) = \left[
\langle \psi | Z_1 | \psi \rangle,
\langle \psi | Z_2 | \psi \rangle,
\dots,
\langle \psi | Z_q | \psi \rangle
\right]
\]

Thus, each quantum circuit produces \(q\) real-valued features.

The full quantum transformation can be written as:
\[
x_i \xrightarrow{\text{reduction}} z_i \xrightarrow{\text{quantum layer}} h_i \in \mathbb{R}^{n_{\text{features}}}
\]

where:
\[
h_i = \bigcup_{k=1}^{c} f(z_i^{(k)})
\]

The final class prediction is computed using a classical linear layer:
\[
\hat{y}_i = W_o h_i + b_o
\]

The model is trained using the cross-entropy loss:
\[
\mathcal{L} = - \sum_i y_i \log(\hat{y}_i)
\]

% Optimization is performed using the Adam optimizer.

The number of qubits per circuit is fixed to \(q = 4\), and the number of circuits is determined by:
\[
c = \frac{n_{\text{features}}}{4}
\]

We evaluate the model under the same variables as the classical baseline.

% \subsubsection{Software dependencies}\label{subsec:software_dep}
\subsection{Experimental Setup}\label{exp}
As an overview, the setup includes: i) Python 3.11.7, ii) PyTorch and \texttt{torchvision} (with an appropriate CUDA build if using GPU)~\cite{paszke2019pytorchimperativestylehighperformance}, iii) For quantum simulation, \texttt{pennylane}~\cite{bergholm2022pennylaneautomaticdifferentiationhybrid} is used, 
% iv) Other Python packages used in the experiment: \texttt{pandas}, \texttt{numpy}, \texttt{matplotlib}, \texttt{scikit-learn}. 
iv) Other Python packages used in the experiment: \texttt{pandas}~\cite{reback2020pandas}, \texttt{numpy}~\cite{harris2020array}, \texttt{matplotlib}~\cite{Hunter:2007}, \texttt{scikit-learn}~\cite{scikit-learn}. 
In this work, NVIDIA H100 80GB HBM3 and Tesla V100-PCIE-16GB GPUs were used.

% CSVM baseline and QSVM
Two experimental settings are considered for the CSVM baseline and QSVM are:
\begin{enumerate}
    \item Feature dimension scaling: the number of PCA components is varied in \{2, 4, 6, 8, 10, 12\}, while the number of samples is fixed at 300.
    \item Sample size scaling: the number of training samples is varied in \{50, 100, 200, 500, 1000\}, while the feature dimension is fixed at 12.
\end{enumerate}
% In both settings, runtime and classification accuracy are recorded. 
The PCA dimension aligns with the quantum model's dimensionality, enabling a controlled comparison between the classical SVM (CSVM) baseline and the quantum model.

% CCNN baseline and QCNN
Two experimental variables are considered for the CCNN baseline and QCNN are:
\begin{enumerate}
    \item Feature dimension: In our experiments, we used feature dimensions, \( d \in \{16, 32, 64\} \), at a sample size of 60,000.
    \item Training sample size: We used \( N \in \{15{,}000, 30{,}000, 60{,}000\} \) to evaluate scaling behavior at feature dimension 16.
\end{enumerate}
% This setup mirrors the experimental configuration of the SVM and QSVM baselines to ensure a controlled comparison between quantum and classical SVMs and neural networks.
Optimization is performed using the Adam optimizer with learning rate \( \eta = 0.001 \). Five epochs are used due to the large computational cost associated with training a quantum neural network.

% \subsubsection{Software dependencies}\label{exp}
% \subsection{Experimental Setup}\label{exp}
% The artifact runs in a Python environment. At a high level, the setup includes: i) Python 3.10. ii) PyTorch and \texttt{torchvision} (with an appropriate CUDA build if using GPU). iii) Hugging Face \texttt{transformers}, \texttt{accelerate}, and \texttt{datasets}. iv) \texttt{sentence-transformers} for MPNet embeddings. v) Common scientific Python packages: \texttt{pandas}, \texttt{numpy}, \texttt{matplotlib}, \texttt{scikit-learn}.
%The README includes a command to create a conda environment and a list of packages to install with \texttt{pip}.

\begin{figure*}[t]   % spans both columns
    \centering
    \includegraphics[width=0.8\textwidth]{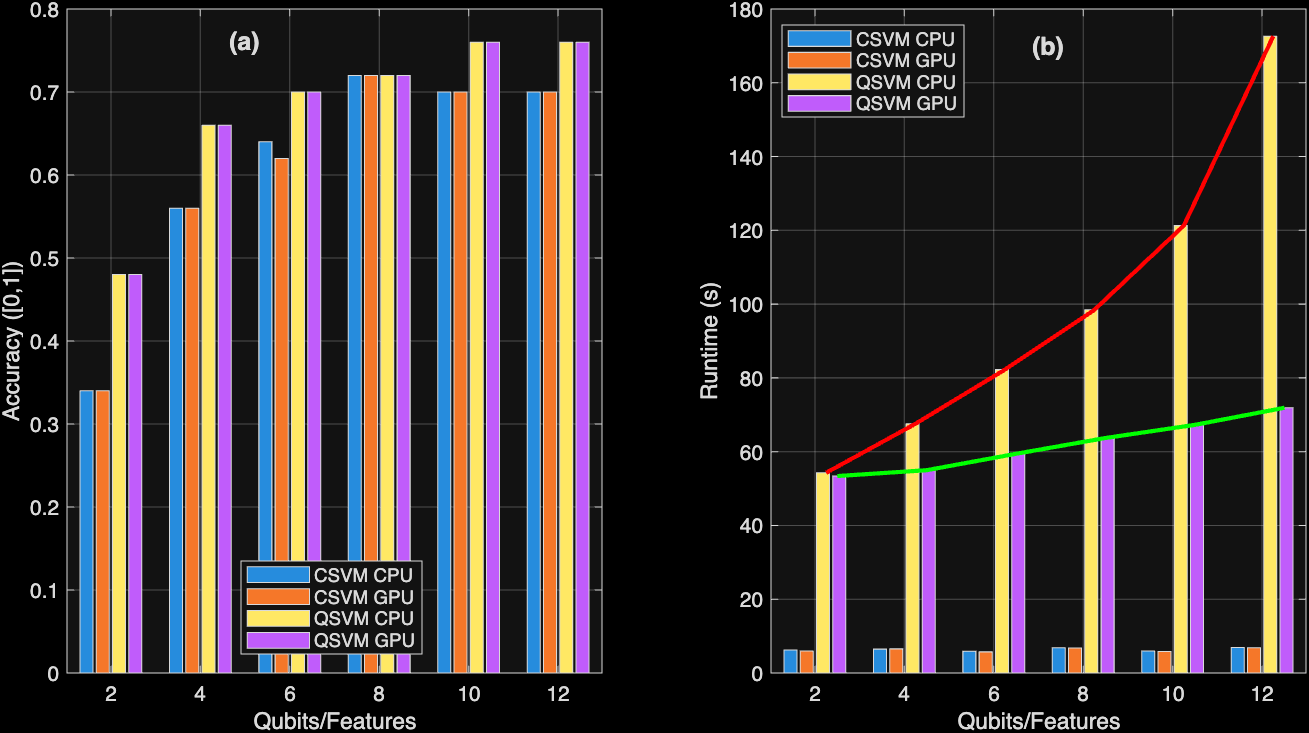}
    \caption{Bar graph plot of (a) accuracy (left) and (b) runtime (right) variation across the CSVM and QSVM with CPU and GPU when varying qubits/features but keeping sample size fixed (i.e., 300).}
    \label{SVM_Acc_RT_FSS}
\end{figure*}

% \begin{figure*}
% \centering
% \subfloat[Accuracy vs. runtime tradeoff of QSVM in GPU]{\includegraphics[width=.45\linewidth]{fig/QSVM_Acc_RT_FSS.png}\label{QSVM_Acc_RT_FSS}}
% % \vfill
% \vspace{20pt}
% \subfloat[Accuracy gap among the best QSVM and CSVM]{\includegraphics[width=.43\linewidth]{fig/Acc_Gap_QSVM_CSVM_FSS.png}\label{Acc_Gap_QSVM_CSVM_FSS}}
% \caption{\textcolor{black}{Effect of varying qubits/features but keeping sample size fixed (i.e., 300)}}
% \label{}
% \end{figure*}

\begin{figure}   % spans both columns
    \centering
    \includegraphics[width=0.45\textwidth]{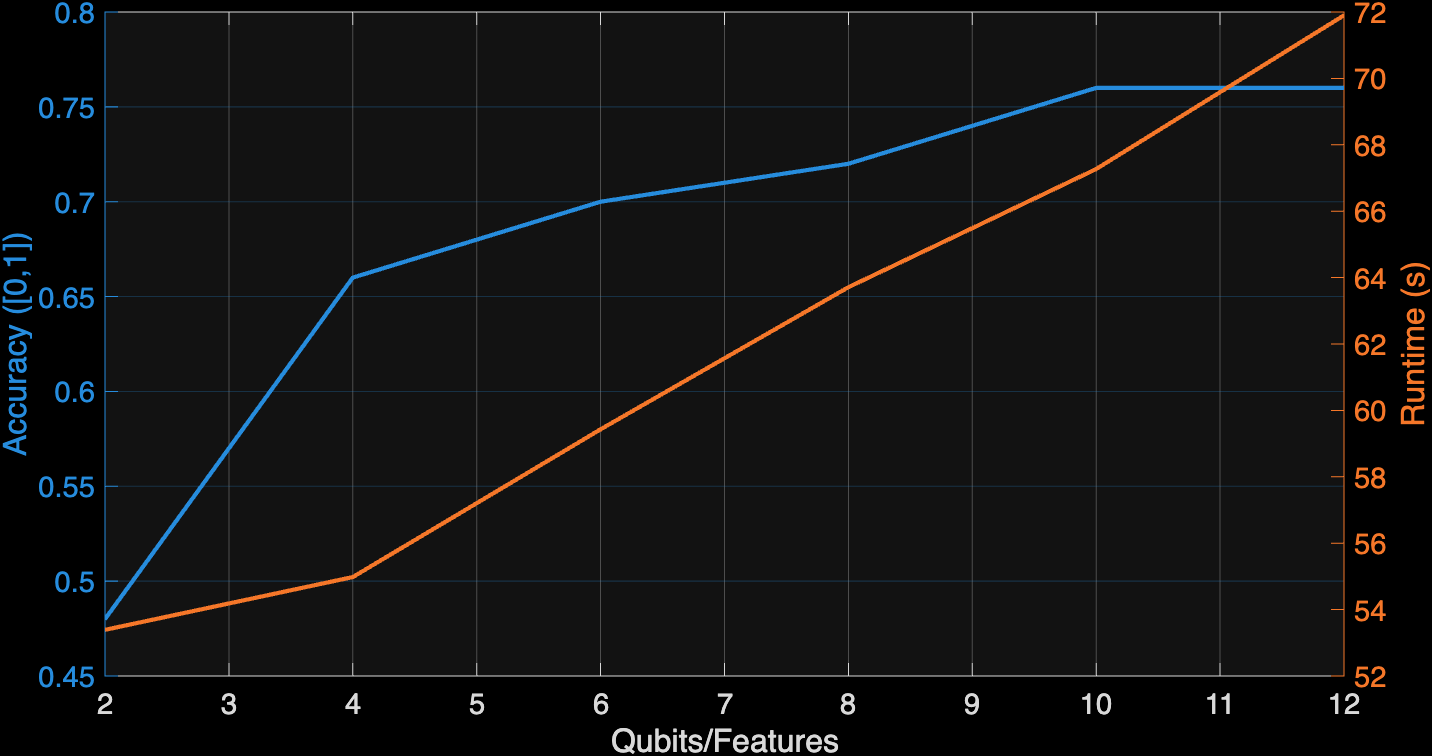}
    \caption{Accuracy vs. runtime tradeoff of QSVM using GPU varying qubits/features but keeping sample size fixed (i.e., 300).}
    \label{QSVM_Acc_RT_FSS}
\end{figure}

\begin{figure*}[t]   % spans both columns
    \centering
    \includegraphics[width=0.8\textwidth]{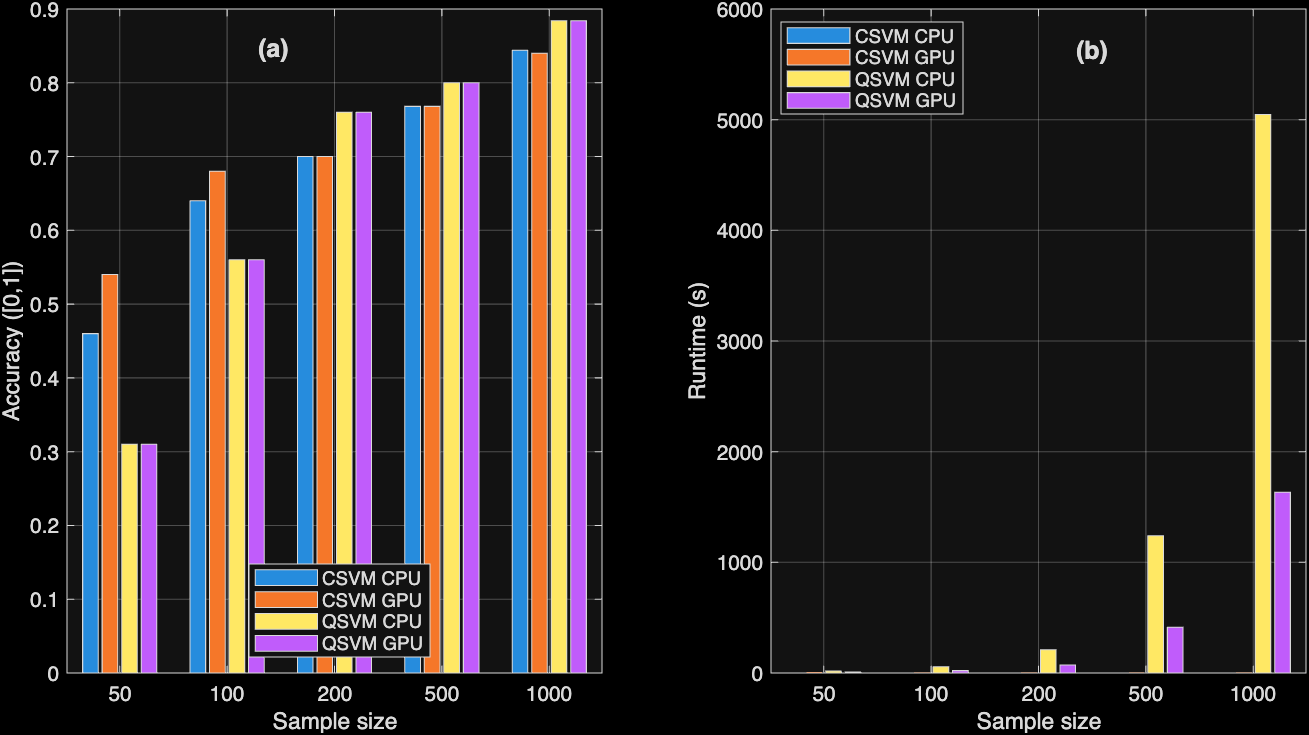}
    % \caption{Accuracy increases steadily with more samples, with quantum lagging behind at smaller sample sizes and surpassing classical at larger ones.}
    \caption{Bar graph plot of (a) accuracy (left) and (b) runtime (right) variation across the CSVM and QSVM with CPU and GPU when varying sample size but keeping qubits/features fixed (i.e., 12).}
    \label{SVM_Acc_RT_FQB}
\end{figure*}
% \section{Results/Analysis}\label{results}
\section{Results}\label{results}
In this section, we analyze the performance of traditional (Section~\ref{tradML}) and deep neural network (Section~\ref{deepML}) models when identifying/classifying the 10 digits in the MNIST dataset using CPU and GPU. To compare the performance of classical and quantum models, we will primarily rely on classification/identification accuracy and runtime. 

\subsection{Traditional Models}\label{tradML}
We will first analyze and compare the performance of Support Vector Machines (SVMs), both the classical SVM (CSVM) and the quantum SVM (QSVM), on CPU and GPU. 

%When Figure~\ref{SVM_Acc_RT_FSS} summarizes our findings when varying qubits/features (i.e., 2, 4, 6, 8, 10, 12), keeping the sample size fixed (i.e., 300), Figure~\ref{SVM_Acc_RT_FQB} presents our findings when varying sample sizes (i.e., 50, 100, 200, 500, 1000), keeping the qubits/features fixed (i.e., 12).

\subsubsection{Effect of Qubit/Feature Count}\label{tradML-FSS}
Figure~\ref{SVM_Acc_RT_FSS} summarizes our findings when varying qubits/features (i.e., 2, 4, 6, 8, 10, 12), keeping the sample size fixed (i.e., 300). 

\noindent{\bf Accuracy Comparison: } 
In Figure~\ref{SVM_Acc_RT_FSS}a, the accuracy plot (left subplot), we observe that, in general, accuracy improves with the increase of qubits/features for both CSVM and QSVM. For a fixed qubit/feature, the quantum SVM (QSVM) usually outperforms the CSVM.

\noindent{\bf Runtime Comparison: } 
In Figure~\ref{SVM_Acc_RT_FSS}b, the runtime plot (right subplot), we observe that the runtime is very low and steady for classical SVMs (CSVMs) compared to quantum SVM (QSVM), which shows an increase with the increase of qubits/features. While the QSVM GPU increases linearly, the QSVM CPU increases exponentially.   

\noindent{\bf Accuracy vs. Runtime Trade-off: }
%\noindent{\bf Optimal Choice of Qubits/Features: } 
From Figure~\ref{SVM_Acc_RT_FSS}, we found QSVM on GPUs to be a better choice, among the four SVM and CPU/GPU combinations, due to its higher accuracy and linearly reasonable runtime. Therefore, for the accuracy vs. runtime trade-off analysis, we chose the QSVM on GPUs. 

In Figure~\ref{QSVM_Acc_RT_FSS}, we observe two crossovers between the dual axis plots of accuracy and runtime at qubit/feature count 2 and 11. However, at qubit/feature count 10, we observe that the accuracy line flattens, while the runtime line continues to sharply rise from qubit/feature count 2 onward. Therefore, qubit/feature count 10 could be an optimal choice to make a trade-off between accuracy and runtime. 

% \begin{figure*}[t]   % spans both columns
%     \centering
%     \includegraphics[width=0.8\textwidth]{fig/Acc_Gap_QSVM_CSVM_FSS.png}
%     \caption{Accuracy gap among the best QSVM and CSVM varying sample size but keeping qubits/features fixed (i.e., 12).}
%     \label{Acc_Gap_QSVM_CSVM_FSS}
% \end{figure*}

% \noindent{\bf Miscellaneous -- Best QSVM vs. CSVM: } 
\noindent{\bf Miscellaneous -- Accuracy Gap Between QSVM \& CSVM: } 
For this analysis, we compare the QSVM and CSVM based on our findings from an accuracy comparison across four combinations of SVM and CPU/GPU (Figure~\ref{SVM_Acc_RT_FSS}a). 
%
%\textcolor{orange}{
% In Figure~\ref{Acc_Gap_QSVM_CSVM_FSS}, we present the accuracy gap between the best QSVM and the best CSVM. 
In the figure, we observe that as the qubit/feature count increases, the accuracy gap between the best QSVM and the best CSVM decreases, with a sharp drop from qubit/feature count 2 to 6, then flattens to 0.06. Therefore, in this setup, after qubit/feature count 6, we observe similar performance between the CSVM and QSVM, with QSVM slightly more accurate. 
%}

\begin{figure*}
\centering
\subfloat[Accuracy vs. runtime tradeoff of QSVM in GPU]{\includegraphics[width=.45\linewidth]{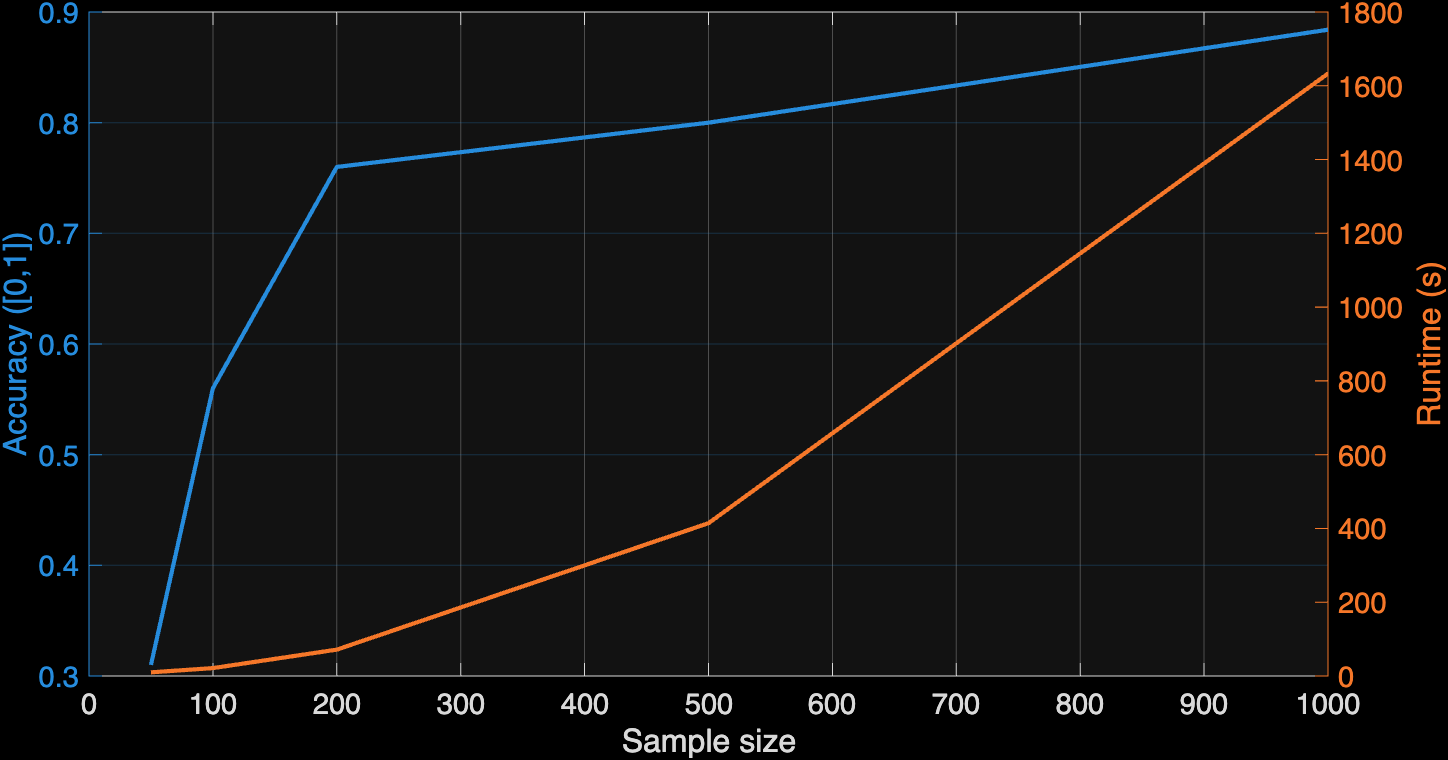}\label{QSVM_Acc_RT_FQB}}
% \vfill
%\vspace{20pt}
\hfill
\subfloat[Accuracy gap among the best QSVM and CSVM (CPU vs. GPU)]{\includegraphics[width=.43\linewidth]{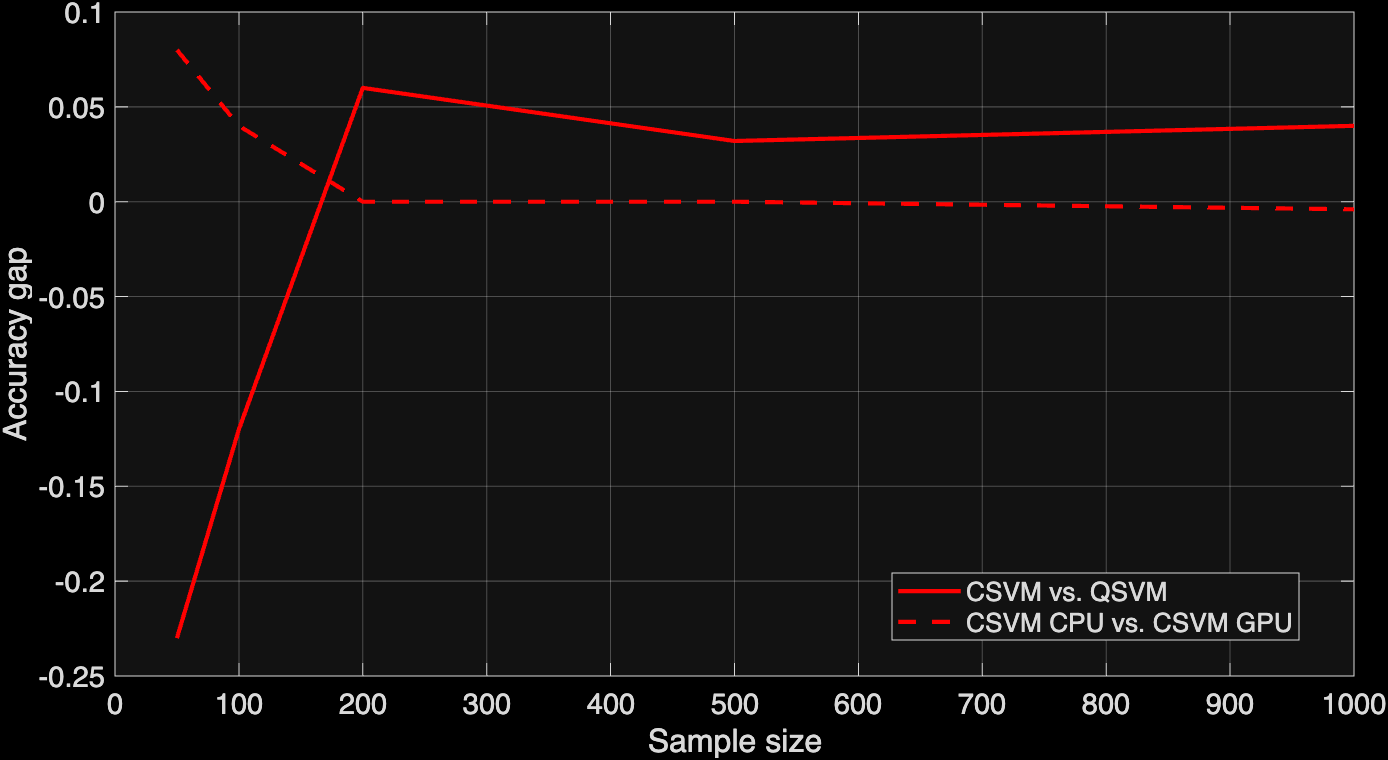}\label{Acc_Gap_QSVM_CSVM_FQB}}
\caption{\textcolor{black}{Effect of varying sample size but keeping qubits/features fixed (i.e., 12)}}
\label{}
\end{figure*}

% \begin{figure*}[t]   % spans both columns
%     \centering
%     \includegraphics[width=0.8\textwidth]{fig/QSVM_Acc_RT_FQB.png}
%     \caption{Accuracy vs. runtime tradeoff of QSVM using GPU varying sample size but keeping qubits/features fixed (i.e., 12).}
%     \label{QSVM_Acc_RT_FQB}
% \end{figure*}

\subsubsection{Effect of Sample Size}\label{tradML-FQB}
Figure~\ref{SVM_Acc_RT_FQB} presents our findings when varying sample sizes (i.e., 50, 100, 200, 500, 1000), keeping the qubits/features fixed (i.e., 12).

\noindent{\bf Accuracy Comparison: } 
In the accuracy plot in Figure~\ref{SVM_Acc_RT_FQB}a (left subplot), we observe that accuracy improves with the increase of sample size for both CSVM and QSVM. At a sample size of 1000, QSVM achieves the highest $\approx$ 0.9 accuracy, and CSVM achieves its highest accuracy of $\approx$ 0.85.

\noindent{\bf Runtime Comparison: } 
In the runtime plot in Figure~\ref{SVM_Acc_RT_FQB}b (right subplot), we observe trends similar to those in Section~\ref{tradML-FSS} as we varied the qubit/feature count, keeping the sample size fixed. The runtime is very low and steady for classical SVMs (CSVMs), varying from 3.23s to 4.28s (CSVM CPU) and from 6.37s to 8.28s (CSVM GPU); whereas the runtime of quantum SVM (QSVM) increases rapidly with increasing sample size, with QSVM CPU runtime increasing faster than QSVM GPU.

\noindent{\bf Accuracy vs. Runtime Trade-off: }
% \noindent{\bf Optimal Choice of Qubits/Features: } 
Similar to the findings in Section~\ref{tradML-FSS}, we found QSVM on GPUs to be a better choice among the four SVMs and CPU/GPU combinations for trade-off analysis, due to its higher accuracy and relatively lower runtime than QSVM on CPU, as found in Figure~\ref{SVM_Acc_RT_FQB}. Therefore, for the accuracy vs. runtime trade-off analysis with varying sample sizes, we chose QSVM on GPUs. 

In Figure~\ref{QSVM_Acc_RT_FQB}, we observe two close proximities between the dual-axis plots of accuracy and runtime at around sample sizes 50 and 1000. While accuracy increases sharply up to a sample size of 200, runtime increases sharply beyond 500. Therefore, a range of [200,500] can be an optimal choice for the sample size (at the fixed qubit/feature count of 12) to balance accuracy and runtime.

% \begin{figure*}[t]   % spans both columns
%     \centering
%     \includegraphics[width=0.8\textwidth]{fig/Acc_Gap_QSVM_CSVM_FQB.png}
%     \caption{Accuracy gap among the best QSVM and CSVM as well as CSVM CPU and CSVM GPU, varying sample size but keeping qubits/features fixed (i.e., 12).}
%     \label{Acc_Gap_QSVM_CSVM_FQB}
% \end{figure*}

\begin{figure*}[t]   % spans both columns
    \centering
    \includegraphics[width=0.8\textwidth]{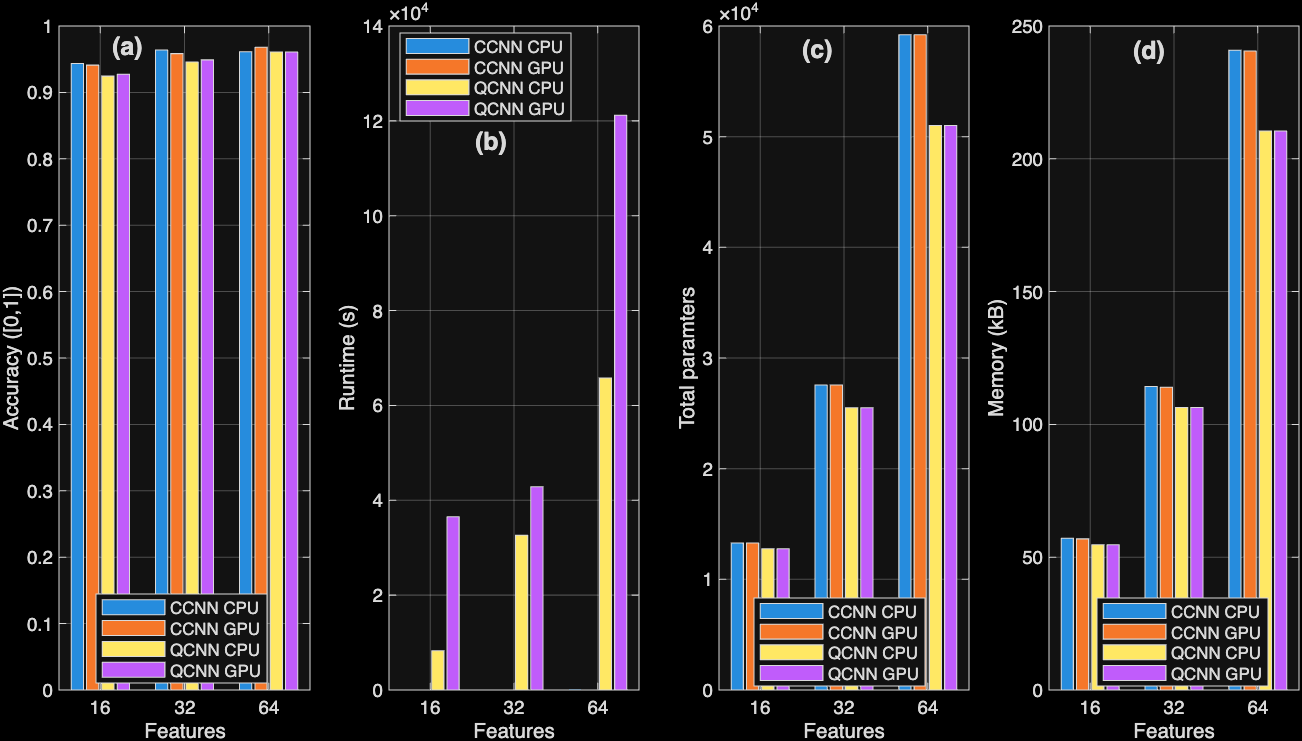}
    \caption{Bar graph plot of (a) accuracy, (b) runtime (s), (c) total parameters, and (d) memory (kB) variation across the CCNN and QCNN with CPU and GPU at epoch 5 when varying features but keeping sample size fixed (i.e., 60000).}
    \label{CNN_Acc_RT_TP_ST_FSS}
\end{figure*}

\noindent{\bf Miscellaneous -- Accuracy Gap Between CPU and GPU versions of CSVM and QSVM: } 
For this analysis, we compare the QSVM and CSVM based on our findings from an accuracy comparison across four combinations of SVM and CPU/GPU (Figure~\ref{SVM_Acc_RT_FQB}a). In that figure, we observed that CSVM accuracy varied across CPUs and GPUs, whereas QSVM achieved similar accuracy across both. Therefore, in addition to comparing the best CSVM and QSVM, we also compare the CSVM on CPU and GPU.

In Figure~\ref{Acc_Gap_QSVM_CSVM_FQB}, we present the accuracy gap between the best CPU and GPU versions of the CSVM as well as the gap between the best CSVM and QSVM. 
First, in Figure~\ref{SVM_Acc_RT_FQB}, we observed that at a lower sample size (i.e., 50), CSVM on the GPU achieves the highest accuracy among the four models, and as the sample size increases, the other models approach the accuracy close to that of CSVM on the GPU. 

In the dashed line in Figure~\ref{Acc_Gap_QSVM_CSVM_FQB}, we observe that as the sample size increases from 50 to 200, the accuracy gap between the CPU and GPU versions of the CSVM decreases faster, and at a sample size of 200, the gap approaches zero and remains steady afterward. Therefore, in this setup with a fixed qubit/feature count of 12, after 200 samples, we would see the same performance across the CPU and GPU versions of the CSVM.

Similarly, though QSVM underperforms CSVM at small sample sizes (negative accuracy gap shown on the solid line in Figure~\ref{Acc_Gap_QSVM_CSVM_FQB}), as the sample size increases, the gap between QSVM and CSVM shrinks, and after 200 samples, QSVM slightly outperforms CSVM (positive gap value shown on the solid line). Therefore, in this setup with a fixed qubit/feature count of 12, after 200 samples, we would see close performance across the CSVM and QSVM, with QSVM being more accurate.

\begin{table}
\caption{Accuracy variation of QCNN and CCNN across different features and epochs for a fixed sample size of 60,000}
\label{Acc-FSS-Epoch}
\begin{center}
\begin{tabular}{l|c|| c c  c c c}
\cline{1-7}
Models   &  Feature & \multicolumn{5}{c}{Accuracy across 5 epochs}    \\
\cline{3-7}
    &  Count &  1   & 2    & 3 & 4 & 5\\
\hline
CCNN    &  16 & 0.9026 & 0.918 & 0.9355 & 0.9391 & 0.9434 \\ 
CPU     &  32 & 0.9348 & 0.9475 & 0.9529 & 0.9526 & 0.964 \\
        &  64 & 0.9432 & 0.9507 & 0.9633 & 0.9675 & 0.9613 \\
\hline
CCNN    &  16 & 0.9094 & 0.9237 & 0.936 & 0.9386 & 0.9411 \\
GPU     &  32 & 0.9348 & 0.9486 & 0.9471 & 0.9579 & 0.9584 \\
        &  64 & 0.9322 & 0.9521 & 0.9606 & 0.9659 & 0.968 \\
\hline
\hline
QCNN    &  16 & 0.8921 & 0.9186 & 0.9105 & 0.9209 & 0.9245 \\ 
CPU     &  32 & 0.9394 & 0.9378 & 0.9454 & 0.9411 & 0.9458 \\ 
        &  64 & 0.9562 & 0.9597 & 0.9627 & 0.9629 & 0.9609 \\
\hline
QCNN    &  16 & 0.8906 & 0.9141 & 0.9191 & 0.9204 & 0.9273 \\ 
GPU     &  32 & 0.9325 & 0.9462 & 0.9487 & 0.9479 & 0.949 \\ 
        &  64 & 0.9572 & 0.9603 & 0.9616 & 0.9607 & 0.9609 \\
\hline
\end{tabular}
\end{center}
\end{table}

\subsection{Deep Neural Network Models}\label{deepML}
In this section, we will analyze and compare the performance of convolutional neural networks (CNNs), both the classical CNN (CCNN) and the quantum CNN (QCNN), on CPU and GPU. For each combination, we vary the epoch count from 1 to 5.

\subsubsection{Effect of Feature Count}\label{deepML-FSS}
Table~\ref{Acc-FSS-Epoch} summarizes the performance of different models across different feature counts (i.e., 16, 32, and 64), varying the epochs while keeping the sample size fixed (i.e., 60,000). In general, we observe that classification/identification accuracy increases with the increase of epochs, both for QCNN and CCNN.  

\noindent{\bf Accuracy Comparison: } 
In Figure~\ref{CNN_Acc_RT_TP_ST_FSS}a, we observe, in general, classification/identification accuracy improves with the increase of feature count. Additionally, while the classical CNN (CCNN) and quantum CNN (QCNN) achieve close accuracy, as the feature count increases, the accuracy gap between the CCNN and QCNN decreases from 0.0161 to 0.0071, consistent with our previous findings (Section~\ref{tradML-FSS}).     

\noindent{\bf Runtime Comparison: } 
In Figure~\ref{CNN_Acc_RT_TP_ST_FSS}b, we observe that the runtime of quantum CNN (QCNN) increases very fast with the increase of feature count, unlike the classical CNN (CCNN), which remains low and steady with the increase of feature count, varying 93.31s -- 172.85s (CCNN CPU) and 94.87s -- 98.22s (CCNN GPU). 
While the QCNN CPU takes 8,255.06s ($\approx$ 2h) to $\approx$ 65,785.89s ($\approx$ 18h), the QCNN GPU takes 36,535.4s ($\approx$ 10h) to  $\approx$ 121168.92s ($\approx$ 33h), with the increase of feature count from 16 to 64. Thereby, QCNN CPU witnesses a higher jump ($\approx$ 88\%) in runtime compared to a $\approx$ 70\% jump in QCNN GPU, with the increase of feature count from 16 to 64, consistent with the exponential versus linear jump in runtime of CPU vs. GPU versions found in Section~\ref{tradML-FSS}.

\noindent{\bf Parameter Count Comparison: } 
In Figure~\ref{CNN_Acc_RT_TP_ST_FSS}c, we observe that, in general, the total number of parameters increases with the increase of feature count across all four combinations. While increasing the feature count from 16 to 64, the accuracy gap between CCNN and QCNN drops from 0.0161 to 0.0071 (i.e., $\approx$ 56\% drop), the difference in total number of parameters between CCNN and QCNN increases from 512 to 8192 (i.e., $\approx$ 94\% more), making the QCNN a good choice at higher feature count, i.e., when parameter count is a major consideration. 

\noindent{\bf Memory Comparison: } 
In Figure~\ref{CNN_Acc_RT_TP_ST_FSS}d, we observe that, in general, the memory requirement increases with the increase of feature count across all four combinations. While increasing the feature count from 16 to 64, the accuracy gap between CCNN and QCNN drops from 0.0161 to 0.0071 (i.e., $\approx$ 56\% drop), the difference in memory requirement between CCNN and QCNN increases from 12705.1kB to 50777.4kB (i.e., $\approx$ 75\% more), making the QCNN a good choice at higher feature count, i.e., when memory is a major consideration. 
This finding is consistent with our finding in Figure~\ref{CNN_Acc_RT_TP_ST_FSS}c. Therefore, QCNN is a better choice than CCNN at higher feature counts and when memory and parameter counts are major considerations.

\begin{table}
\caption{Accuracy variation of QCNN and CCNN across different sample sizes and epochs for feature count 16}
\label{Acc-FQB-Epoch}
\begin{center}
\begin{tabular}{l|c|| c c  c c c}
\cline{1-7}
Models   &  Sample & \multicolumn{5}{c}{Accuracy across 5 epochs}    \\
\cline{3-7}
    & Size &  1   & 2    & 3 & 4 & 5\\
\hline
CCNN    &  15000 & 0.8842 & 0.9065 & 0.9104 & 0.9163 & 0.9229 \\ 
CPU     &  30000 & 0.8999 & 0.9118 & 0.9116 & 0.9265 & 0.9311 \\ 
        &  60000 & 0.9026 & 0.918 & 0.9355 & 0.9391 & 0.9434 \\  
\hline
CCNN    &  15000 & 0.8596 & 0.8909 & 0.9024 & 0.9131 & 0.9203 \\ 
GPU     &  30000 & 0.8927 & 0.9117 & 0.9227 & 0.9272 & 0.9322 \\  
        &  60000 & 0.9157 & 0.9354 & 0.9385 & 0.9462 & 0.9442 \\  
\hline
\hline
QCNN    &  15000 & 0.8539 & 0.8808 & 0.8911 & 0.899 & 0.9017 \\ 
CPU     &  30000 & 0.8868 & 0.9046 & 0.9056 & 0.9058 & 0.907 \\
        &  60000 & 0.8921 & 0.9186 & 0.9105 & 0.9209 & 0.9245 \\
\hline
QCNN    &  15000 & 0.8539 & 0.8808 & 0.8911 & 0.899 & 0.9017 \\ 
GPU     &  30000 & 0.8646 & 0.8854 & 0.8966 & 0.9033 & 0.9109 \\
        &  60000 & 0.8921 & 0.9186 & 0.9105 & 0.9209 & 0.9245 \\
\hline
\end{tabular}
\end{center}
\end{table}

\begin{figure*}[t]   % spans both columns
    \centering
    \includegraphics[width=0.8\textwidth]{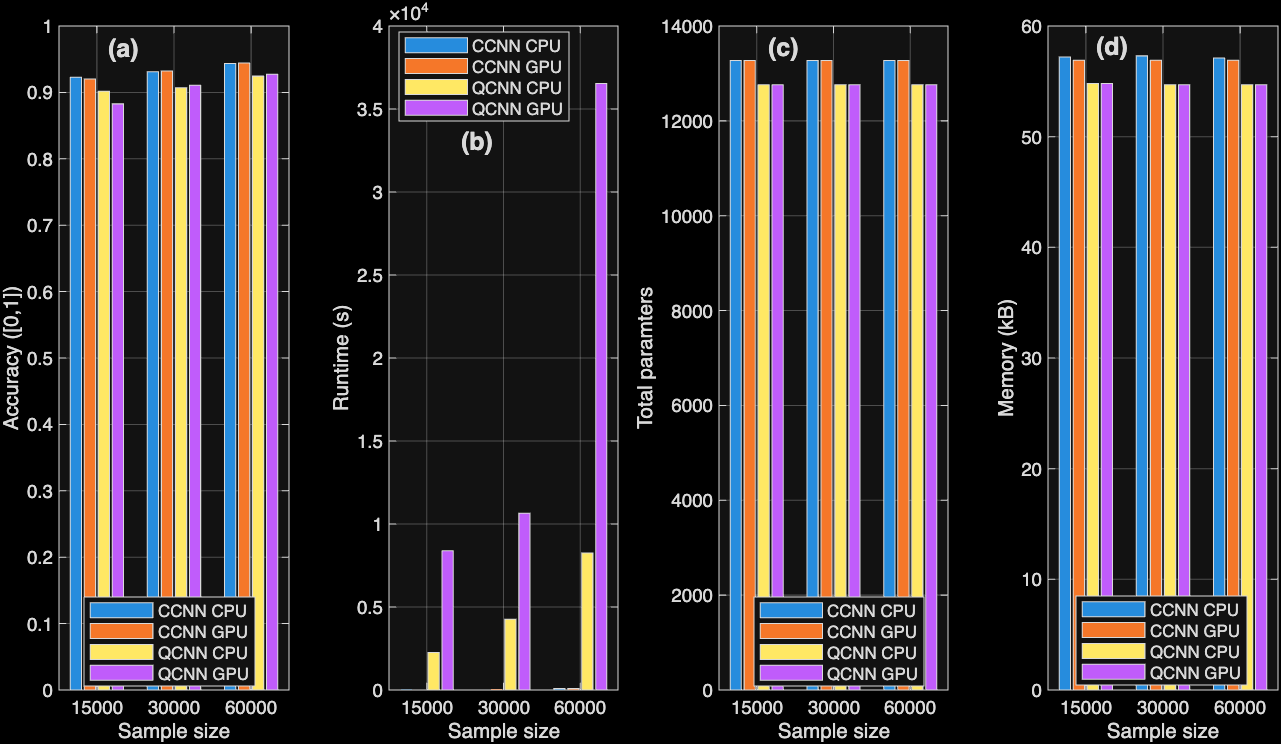}
    \caption{Bar graph plot of (a) accuracy, (b) runtime (s), (c) total parameters, and (d) memory (kB) variation across the CCNN and QCNN with CPU and GPU at epoch 5 when varying sample size but keeping features fixed (i.e., 16).}
    \label{CNN_Acc_RT_TP_ST_FQB}
\end{figure*}

\subsubsection{Effect of Sample Size}\label{deepML-FQB}
Table~\ref{Acc-FQB-Epoch} summarizes the performance of different models across different sample sizes (i.e., 15,000, 30,000, and 60,000), varying the epochs while keeping the feature count fixed (i.e., 16). In general, we observe that classification/identification accuracy increases with increasing epochs, both for QCNN and CCNN. 

\noindent{\bf Accuracy Comparison: } 
In Figure~\ref{CNN_Acc_RT_TP_ST_FQB}a, we observe, in general, classification/identification accuracy improves with the increase of sample size at a fixed feature count. Additionally, while the CCNN and QCNN achieve close accuracy, as the sample size increases from 15,000 to 60,000, the accuracy gap between them decreases from 0.0212 to 0.0169, consistent with our previous findings (Section~\ref{tradML-FQB}).  

\noindent{\bf Runtime Comparison: } 
In Figure~\ref{CNN_Acc_RT_TP_ST_FQB}b, we observe that the runtime of quantum CNN (QCNN) increases very fast with the increase of sample size, unlike the classical CNN (CCNN), which remains low and steady with the increase of feature count, varying 41.19s -- 93.31s (CCNN CPU) and 33.08s -- 94.27s (CCNN GPU). 
While the QCNN CPU takes 2,261.61s ($\approx$ 38 minutes) to $\approx$ 8,255.06s ($\approx$ 2h), the QCNN GPU takes 8,389.86s ($\approx$ 2h) to  $\approx$ 36,535.4s ($\approx$ 10h), with the increase of feature count from 16 to 64. Thereby, QCNN CPU witnesses a similar jump ($\approx$ 73\%) in runtime as QCNN GPU ($\approx$ 77\%), with the increase in sample size from 15,000 to 60,000.

\noindent{\bf Parameter Count Comparison: } 
In Figure~\ref{CNN_Acc_RT_TP_ST_FQB}c, we observe that parameter count remains unchanged with the increase of sample size across all four combinations, while QCNN requires 512 fewer parameters than the CCNN to achieve similar performance, i.e., accuracy. However, in Figure~\ref {CNN_Acc_RT_TP_ST_FQB}a, we found that the accuracy gap between QCNN and CCNN reduces with the increase of sample size. Therefore, the QCNN can be a good choice for larger sample sizes. 

\noindent{\bf Memory Comparison: } 
Similar to our findings in Figure~\ref{CNN_Acc_RT_TP_ST_FQB}c, we observe in Figure~\ref{CNN_Acc_RT_TP_ST_FQB}d that the memory requirement remains unchanged with increasing sample size across all four combinations, while QCNN requires 12705.1kB less memory than CCNN to achieve similar accuracy. However, in Figure~\ref {CNN_Acc_RT_TP_ST_FQB}a, we found that the accuracy gap between QCNN and CCNN reduces with the increase of sample size. Therefore, the QCNN can be a memory-efficient choice at larger sample sizes.
This finding is consistent with our finding in Figure~\ref{CNN_Acc_RT_TP_ST_FQB}c. Therefore, QCNN is a better choice than CCNN at higher sample sizes and when memory and parameter counts are major considerations.

\section{Conclusion}\label{sec:concl}

This study benchmarks classical and quantum machine learning models, i.e., SVMs and CNNs, across accuracy, runtime, parameter efficiency, and memory on the MNIST dataset. For the SVM-based models, QSVM consistently outperforms CSVM in accuracy, particularly at larger sample sizes (reaching $\sim$ 0.9 vs. $\sim$ 0.85 at 1,000 samples), though this comes at a higher computational cost. As QSVM runtime scales exponentially on CPU and linearly on GPU, GPU execution is the only practical choice for QSVMs. A feature count of 10 qubits and a sample size of 200 -- 500 emerge as reasonable operating points that balance accuracy and runtime. For the neural network models, CCNN and QCNN achieve comparable classification accuracy (both exceeding 0.96 at 64 features and 60,000 samples), but QCNN offers better parameter and memory efficiency at higher feature counts, while suffering from longer runtimes. Across both model families, quantum ML models outperform the classical ML models more as feature count or sample size increases, suggesting that quantum models offer the greatest relative advantage in high-data or high-dimensionality regimes.

Several promising avenues remain unexplored. First, experiments here were limited to simulated quantum hardware via PennyLane; evaluating these models on actual quantum hardware would expose the impact of noise, decoherence, and gate errors on both accuracy and runtime, factors absent from simulation. Second, scaling beyond the small qubit counts used here (up to 12 for QSVM, 4 per circuit for QCNN) is critical, as quantum advantage may emerge at higher dimensionalities that current simulators make prohibitively expensive to simulate. Third, the dataset was restricted to MNIST, a relatively simple benchmark; testing on more complex image datasets (e.g., CIFAR-10 or ImageNet subsets) would better stress-test the generalization capacity of quantum models. Fourth, exploring more expressive quantum circuit architectures, such as deeper variational circuits, alternative encoding strategies beyond angle embedding, or error-mitigation techniques, could improve accuracy while potentially reducing the runtime gap with classical counterparts. Finally, a systematic study of hybrid architectures that selectively apply quantum layers only where they provide measurable benefit could offer a pragmatic path toward quantum-classical co-design in real-world computer vision pipelines. 

%Our future plan is to extend findings from this work to other sectors of computer vision and image recognition, including real-time traffic and road monitoring, toward an AI-assisted smart transportation system. 
Looking ahead, this work directly supports the mission of the Alabama Transportation Institute (ATI) at The University of Alabama, to facilitate world-class, interdisciplinary transportation research that serves the state of Alabama and beyond, as well as the broader mandate of the National Center for Transportation Cybersecurity and Resiliency (TraCR), headquartered at Clemson University, which explicitly identifies artificial intelligence and quantum computing as transformative technologies for strengthening the cybersecurity and resiliency of the nation's transportation systems. ATI's strategic commitment to \textit{ACES\textsuperscript{2} Mobility}, mobility that is \textbf{A}utomated, \textbf{C}onnected, \textbf{E}lectric, \textbf{S}hared, and \textbf{S}afe, demands robust, real-time computer vision capabilities for autonomous perception, object detection, traffic monitoring, and anomaly identification in connected and autonomous vehicles (CAVs) and intelligent transportation infrastructure. The benchmarking insights derived in this work, particularly the identification of practical operating regimes in terms of qubit count, sample size, and hardware environment where quantum models deliver measurable gains in accuracy and resource efficiency over classical counterparts, directly inform the design of next-generation, computationally efficient perception pipelines for such transportation applications. Specifically, the finding that QCNN requires $\sim$94\% fewer parameters and $\sim$75\% less memory than CCNN at higher feature counts, while achieving comparable accuracy, is especially relevant for resource-constrained onboard computing in CAVs and edge-deployed transportation sensors. Furthermore, the superior parameter efficiency of quantum models aligns with TraCR's goal of developing adaptive and resilient cyber-physical transportation systems, where lightweight yet accurate vision models can improve robustness against adversarial perturbations and reduce the attack surface of onboard AI inference systems. Our future plan is therefore to extend and apply the findings from this work directly to real-world transportation vision tasks supported by ATI and TraCR, including real-time traffic and road monitoring, pedestrian detection, vehicle classification at roadway intersections, and anomaly detection in connected infrastructure, working toward a safer, more efficient, and quantum-ready AI-assisted smart transportation ecosystem for Alabama and the nation.

\section*{Acknowledgement} 
%The authors would like to thank the reviewers for their constructive feedback to improve the manuscript. 
%The work has been partially funded by anonymous XX and YY grants.
The author(s) gratefully acknowledge support from the Alabama Transportation Institute at The University of Alabama. This material is based upon the work supported by the National Center for Transportation Cybersecurity and Resiliency (TraCR) headquartered at Clemson University, Clemson, South Carolina, USA. Any opinions, findings, conclusions, and recommendations expressed in this material are those of the author(s) and do not necessarily reflect the views of TraCR, and the U.S. Government assumes no liability for the contents or use thereof.

\ignore{====================
\section{Ease of Use}

\subsection{Maintaining the Integrity of the Specifications}

The IEEEtran class file is used to format your paper and style the text. All margins, 
column widths, line spaces, and text fonts are prescribed; please do not 
alter them. You may note peculiarities. For example, the head margin
measures proportionately more than is customary. This measurement 
and others are deliberate, using specifications that anticipate your paper 
as one part of the entire proceedings, and not as an independent document. 
Please do not revise any of the current designations.

\section{Prepare Your Paper Before Styling}
Before you begin to format your paper, first write and save the content as a 
separate text file. Complete all content and organizational editing before 
formatting. Please note sections \ref{AA}--\ref{SCM} below for more information on 
proofreading, spelling and grammar.

Keep your text and graphic files separate until after the text has been 
formatted and styled. Do not number text heads---{\LaTeX} will do that 
for you.

\subsection{Abbreviations and Acronyms}\label{AA}
Define abbreviations and acronyms the first time they are used in the text, 
even after they have been defined in the abstract. Abbreviations such as 
IEEE, SI, MKS, CGS, ac, dc, and rms do not have to be defined. Do not use 
abbreviations in the title or heads unless they are unavoidable.

\subsection{Units}
\begin{itemize}
\item Use either SI (MKS) or CGS as primary units. (SI units are encouraged.) English units may be used as secondary units (in parentheses). An exception would be the use of English units as identifiers in trade, such as ``3.5-inch disk drive''.
\item Avoid combining SI and CGS units, such as current in amperes and magnetic field in oersteds. This often leads to confusion because equations do not balance dimensionally. If you must use mixed units, clearly state the units for each quantity that you use in an equation.
\item Do not mix complete spellings and abbreviations of units: ``Wb/m\textsuperscript{2}'' or ``webers per square meter'', not ``webers/m\textsuperscript{2}''. Spell out units when they appear in text: ``. . . a few henries'', not ``. . . a few H''.
\item Use a zero before decimal points: ``0.25'', not ``.25''. Use ``cm\textsuperscript{3}'', not ``cc''.)
\end{itemize}

\subsection{Equations}
Number equations consecutively. To make your 
equations more compact, you may use the solidus (~/~), the exp function, or 
appropriate exponents. Italicize Roman symbols for quantities and variables, 
but not Greek symbols. Use a long dash rather than a hyphen for a minus 
sign. Punctuate equations with commas or periods when they are part of a 
sentence, as in:
\begin{equation}
a+b=\gamma\label{eq}
\end{equation}

Be sure that the 
symbols in your equation have been defined before or immediately following 
the equation. Use ``\eqref{eq}'', not ``Eq.~\eqref{eq}'' or ``equation \eqref{eq}'', except at 
the beginning of a sentence: ``Equation \eqref{eq} is . . .''

\subsection{\LaTeX-Specific Advice}

Please use ``soft'' (e.g., \verb|\eqref{Eq}|) cross references instead
of ``hard'' references (e.g., \verb|(1)|). That will make it possible
to combine sections, add equations, or change the order of figures or
citations without having to go through the file line by line.

Please don't use the \verb|{eqnarray}| equation environment. Use
\verb|{align}| or \verb|{IEEEeqnarray}| instead. The \verb|{eqnarray}|
environment leaves unsightly spaces around relation symbols.

Please note that the \verb|{subequations}| environment in {\LaTeX}
will increment the main equation counter even when there are no
equation numbers displayed. If you forget that, you might write an
article in which the equation numbers skip from (17) to (20), causing
the copy editors to wonder if you've discovered a new method of
counting.

{\BibTeX} does not work by magic. It doesn't get the bibliographic
data from thin air but from .bib files. If you use {\BibTeX} to produce a
bibliography you must send the .bib files. 

{\LaTeX} can't read your mind. If you assign the same label to a
subsubsection and a table, you might find that Table I has been cross
referenced as Table IV-B3. 

{\LaTeX} does not have precognitive abilities. If you put a
\verb|\label| command before the command that updates the counter it's
supposed to be using, the label will pick up the last counter to be
cross referenced instead. In particular, a \verb|\label| command
should not go before the caption of a figure or a table.

Do not use \verb|\nonumber| inside the \verb|{array}| environment. It
will not stop equation numbers inside \verb|{array}| (there won't be
any anyway) and it might stop a wanted equation number in the
surrounding equation.

\subsection{Some Common Mistakes}\label{SCM}
\begin{itemize}
\item The word ``data'' is plural, not singular.
\item The subscript for the permeability of vacuum $\mu_{0}$, and other common scientific constants, is zero with subscript formatting, not a lowercase letter ``o''.
\item In American English, commas, semicolons, periods, question and exclamation marks are located within quotation marks only when a complete thought or name is cited, such as a title or full quotation. When quotation marks are used, instead of a bold or italic typeface, to highlight a word or phrase, punctuation should appear outside of the quotation marks. A parenthetical phrase or statement at the end of a sentence is punctuated outside of the closing parenthesis (like this). (A parenthetical sentence is punctuated within the parentheses.)
\item A graph within a graph is an ``inset'', not an ``insert''. The word alternatively is preferred to the word ``alternately'' (unless you really mean something that alternates).
\item Do not use the word ``essentially'' to mean ``approximately'' or ``effectively''.
\item In your paper title, if the words ``that uses'' can accurately replace the word ``using'', capitalize the ``u''; if not, keep using lower-cased.
\item Be aware of the different meanings of the homophones ``affect'' and ``effect'', ``complement'' and ``compliment'', ``discreet'' and ``discrete'', ``principal'' and ``principle''.
\item Do not confuse ``imply'' and ``infer''.
\item The prefix ``non'' is not a word; it should be joined to the word it modifies, usually without a hyphen.
\item There is no period after the ``et'' in the Latin abbreviation ``et al.''.
\item The abbreviation ``i.e.'' means ``that is'', and the abbreviation ``e.g.'' means ``for example''.
\end{itemize}
An excellent style manual for science writers is \cite{b7}.

\subsection{Authors and Affiliations}
\textbf{The class file is designed for, but not limited to, six authors.} A 
minimum of one author is required for all conference articles. Author names 
should be listed starting from left to right and then moving down to the 
next line. This is the author sequence that will be used in future citations 
and by indexing services. Names should not be listed in columns nor group by 
affiliation. Please keep your affiliations as succinct as possible (for 
example, do not differentiate among departments of the same organization).

\subsection{Identify the Headings}
Headings, or heads, are organizational devices that guide the reader through 
your paper. There are two types: component heads and text heads.

Component heads identify the different components of your paper and are not 
topically subordinate to each other. Examples include Acknowledgments and 
References and, for these, the correct style to use is ``Heading 5''. Use 
``figure caption'' for your Figure captions, and ``table head'' for your 
table title. Run-in heads, such as ``Abstract'', will require you to apply a 
style (in this case, italic) in addition to the style provided by the drop 
down menu to differentiate the head from the text.

Text heads organize the topics on a relational, hierarchical basis. For 
example, the paper title is the primary text head because all subsequent 
material relates and elaborates on this one topic. If there are two or more 
sub-topics, the next level head (uppercase Roman numerals) should be used 
and, conversely, if there are not at least two sub-topics, then no subheads 
should be introduced.

\subsection{Figures and Tables}
\paragraph{Positioning Figures and Tables} Place figures and tables at the top and 
bottom of columns. Avoid placing them in the middle of columns. Large 
figures and tables may span across both columns. Figure captions should be 
below the figures; table heads should appear above the tables. Insert 
figures and tables after they are cited in the text. Use the abbreviation 
``Fig.~\ref{fig}'', even at the beginning of a sentence.

\begin{table}[htbp]
\caption{Table Type Styles}
\begin{center}
\begin{tabular}{|c|c|c|c|}
\hline
\textbf{Table}&\multicolumn{3}{|c|}{\textbf{Table Column Head}} \\
\cline{2-4} 
\textbf{Head} & \textbf{\textit{Table column subhead}}& \textbf{\textit{Subhead}}& \textbf{\textit{Subhead}} \\
\hline
copy& More table copy$^{\mathrm{a}}$& &  \\
\hline
\multicolumn{4}{l}{$^{\mathrm{a}}$Sample of a Table footnote.}
\end{tabular}
\label{tab1}
\end{center}
\end{table}

\begin{figure}[htbp]
\centerline{\includegraphics{fig1.png}}
\caption{Example of a figure caption.}
\label{fig}
\end{figure}

Figure Labels: Use 8 point Times New Roman for Figure labels. Use words 
rather than symbols or abbreviations when writing Figure axis labels to 
avoid confusing the reader. As an example, write the quantity 
``Magnetization'', or ``Magnetization, M'', not just ``M''. If including 
units in the label, present them within parentheses. Do not label axes only 
with units. In the example, write ``Magnetization (A/m)'' or ``Magnetization 
\{A[m(1)]\}'', not just ``A/m''. Do not label axes with a ratio of 
quantities and units. For example, write ``Temperature (K)'', not 
``Temperature/K''.

\section*{Acknowledgment}

The preferred spelling of the word ``acknowledgment'' in America is without 
an ``e'' after the ``g''. Avoid the stilted expression ``one of us (R. B. 
G.) thanks $\ldots$''. Instead, try ``R. B. G. thanks$\ldots$''. Put sponsor 
acknowledgments in the unnumbered footnote on the first page.

\section*{References}

Please number citations consecutively within brackets \cite{b1}. The 
sentence punctuation follows the bracket \cite{b2}. Refer simply to the reference 
number, as in \cite{b3}---do not use ``Ref. \cite{b3}'' or ``reference \cite{b3}'' except at 
the beginning of a sentence: ``Reference \cite{b3} was the first $\ldots$''

Number footnotes separately in superscripts. Place the actual footnote at 
the bottom of the column in which it was cited. Do not put footnotes in the 
abstract or reference list. Use letters for table footnotes.

Unless there are six authors or more give all authors' names; do not use 
``et al.''. Papers that have not been published, even if they have been 
submitted for publication, should be cited as ``unpublished'' \cite{b4}. Papers 
that have been accepted for publication should be cited as ``in press'' \cite{b5}. 
Capitalize only the first word in a paper title, except for proper nouns and 
element symbols.

For papers published in translation journals, please give the English 
citation first, followed by the original foreign-language citation \cite{b6}.

\vspace{12pt}
\color{red}
IEEE conference templates contain guidance text for composing and formatting conference papers. Please ensure that all template text is removed from your conference paper prior to submission to the conference. Failure to remove the template text from your paper may result in your paper not being published.
====================}

\bibliographystyle{ieeetr} % Or abbrvnat, which are compatible with the NeurIPS style
% \notice{*}\bibliography{ref}
\bibliography{references}

\begin{thebibliography}{00}
\bibitem{b1} G. Eason, B. Noble, and I. N. Sneddon, ``On certain integrals of Lipschitz-Hankel type involving products of Bessel functions,'' Phil. Trans. Roy. Soc. London, vol. A247, pp. 529--551, April 1955.
\bibitem{b2} J. Clerk Maxwell, A Treatise on Electricity and Magnetism, 3rd ed., vol. 2. Oxford: Clarendon, 1892, pp.68--73.
\bibitem{b3} I. S. Jacobs and C. P. Bean, ``Fine particles, thin films and exchange anisotropy,'' in Magnetism, vol. III, G. T. Rado and H. Suhl, Eds. New York: Academic, 1963, pp. 271--350.
\bibitem{b4} K. Elissa, ``Title of paper if known,'' unpublished.
\bibitem{b5} R. Nicole, ``Title of paper with only first word capitalized,'' J. Name Stand. Abbrev., in press.
\bibitem{b6} Y. Yorozu, M. Hirano, K. Oka, and Y. Tagawa, ``Electron spectroscopy studies on magneto-optical media and plastic substrate interface,'' IEEE Transl. J. Magn. Japan, vol. 2, pp. 740--741, August 1987 [Digests 9th Annual Conf. Magnetics Japan, p. 301, 1982].
\bibitem{b7} M. Young, The Technical Writer's Handbook. Mill Valley, CA: University Science, 1989.
\end{thebibliography}

\end{document}